\documentclass[twocolumn,cleanfoot, 10pt]{asme2e}
\usepackage{textcase}

\usepackage{setspace}
\usepackage{enumitem}
\usepackage{amsmath}    
\usepackage{amsfonts}   
\usepackage{amssymb}   
\usepackage{graphicx}
\usepackage{float}
\usepackage{subcaption}
\usepackage{stfloats}
\usepackage{xcolor}
\usepackage{soul}
\usepackage[normalem]{ulem}

\makeatletter
\renewcommand\subsubsection{\@startsection{subsubsection}{3}{0pt}%
  {-1.25ex \@plus -1ex \@minus -.2ex}%
  {0.5ex \@plus .2ex}%
  {\normalfont\normalsize\bfseries}}
\makeatother

\confshortname{IDETC-CIE2026\\[3pt]}
\conffullname{the ASME 2026 \\
International Design Engineering Technical Conferences and\\
              Computers and Information in Engineering Conference\\}

\confdate{August 23-26}
\confyear{2026}
\confcity{Houston}
\confcountry{TX}

\papernum{DETC2026-194768}

\title{\NoCaseChange{ShowerFlex: Achieving Pseudo-Static Balancing in a Continuum Shower Hose}}

\author{Zhiyu Ren
    \affiliation{Department of Industrial and Systems Engineering\\
	University of Illinois, Urbana-Champaign\\
	Urbana, Illinois 61801\\
    Email: zhiyur2@illinois.edu
    }	
}

\author{Girish Krishnan\thanks{Address all correspondence to this author.}
    \affiliation{Department of Industrial and Systems Engineering\\
	University of Illinois, Urbana-Champaign\\
	Urbana, Illinois 61801\\
	Email: gkrishna@illinois.edu
    }
}

\begin{document}

\maketitle    

\begin{abstract}
{\it With the global population rapidly aging, maintaining independence in Activities of Daily Living (ADLs)—particularly bathing or showering—has become a critical challenge. Older adults who sit while showering currently have limited options, often relying on rigid overhead showerheads or flexible hand-held hoses that require continuous gripping and precarious user maneuvers, increasing the risk of falls. To address this need, this paper introduces a highly articulated, pseudo-static balanced continuum mechanism designed specifically for accessible bathing assistance. The proposed mechanism features a modular continuum architecture composed of friction-locked ball-and-socket joints (loc-line), seamlessly integrated with a retractable spring-loaded reel for effective gravity compensation. This hybrid design achieves an intuitive “Push-and-Stay” interaction logic, maintaining an approximate static equilibrium with ergonomically low actuation force, without any external electronic power, while significantly minimizing the force needed for repositioning. Theoretically, we established a recursive forward kinematics framework to construct the geometric model of the structure, coupled with a comprehensive static equilibrium analysis to evaluate the system's holding capacity across its 51-DOF structure. Numerical simulations utilizing Monte Carlo methods validate the system's morphological adaptability and stability at various extreme positions within a standard bath space. Physical experiments further validate the prototype's real-world performance, confirming its intrinsic static stability against gravity and ensuring that the actuation force remains well within the ergonomic capabilities of older adults. Ultimately, this research provides a low-cost, intrinsically safe showerhead design that reduces physical strain, restoring dignity and autonomy in personal hygiene.}
\end{abstract}


\begin{figure}[H]
    \centering
    \includegraphics[height=6cm]{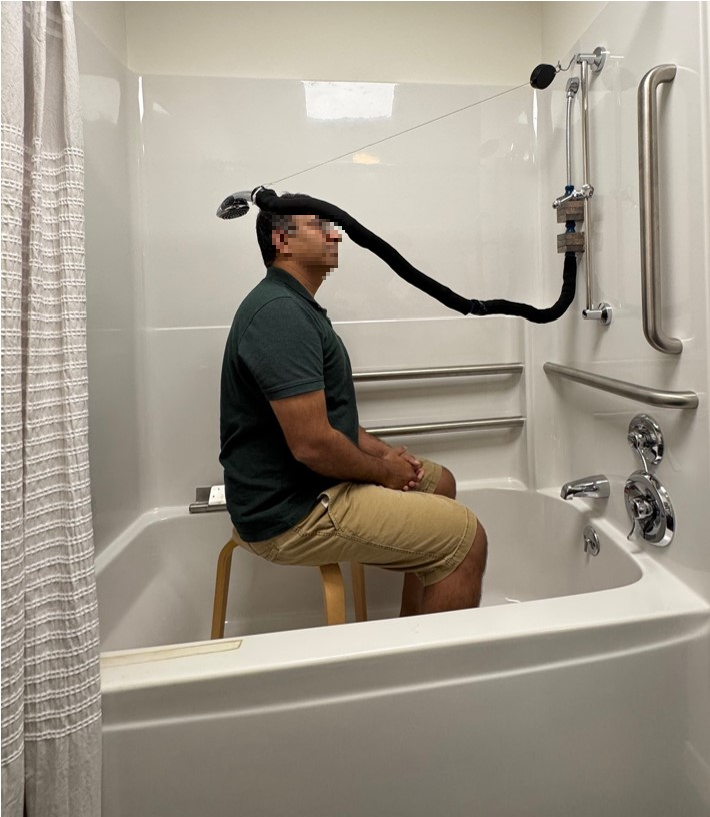}
    \caption{System overview of the proposed ShowerFlex bathing assistant}
    \label{fig:introduction}
\end{figure}
\section{INTRODUCTION}
The global community is rapidly aging. According to the latest projections from the World Health Organization (2025), the global population of individuals aged 60 and over is expected to double, reaching 2.1 billion by 2050~\cite{WHO2025AgeingHealth}~\cite{jugran2025too}~\cite{ONS-2024-visibility-older}. Within this demographic shift, maintaining independence in “Activities of Daily Living” (ADLs), and in particular maintaining personal hygiene through routine showering or bathing stands as a critical determinant of healthy aging. Independent bathing is often the most important physiological threshold where functional decline becomes apparent~\cite{Edemekong2025ADL}~\cite{bathingdecline}. The inability to perform this intimate task not only poses physical health risks but also precipitates a severe decline in psychological well-being, often leading to a loss of dignity and a heightened sense of institutional dependency~\cite{katz1983assessing}~\cite{dignity}.
For older adults who use wheelchairs or have trouble moving their legs, staying independent in the shower is difficult for two main reasons. First, moving from a wheelchair into the shower is physically demanding. To make this easier, homes often need structural changes, such as replacing traditional tubs with “walk-in" showers that have no curb or a very low entry point for a wheelchair to roll over. Second, for those who shower while sitting, ensuring that the shower reaches certain parts of the body like the back, head, and feet is often impossible ~\cite{khomami2021survey}. Reaching the water temperature and flow controls can also be a struggle. This paper focuses on that second challenge: making it easier for older adults to reach their body and operate shower controls comfortably while seated.

Despite these well-documented challenges, shower hardware has remained largely stagnant for decades. Traditional fixed showerheads are reliable, but their rigid design cannot adapt to the human body’s shape. This forces users into risky physical movements just to reach their back or feet ~\cite{khomami2021survey}. While adjustable showerheads from brands like Waterpik or Delta exist, they are designed for standing adults of different heights. They are ineffective for those who shower while seated, as the user typically sits three to four feet away from the water source. Handheld showerheads with flexible hoses are a common alternative, but they require the user to constantly hold and aim them. For frail older adults, this is a major safety risk: if one hand is busy holding a showerhead, they cannot use both hands to grip safety bars, significantly increasing the risk of a fall ~\cite{werner2022evaluating}.

To address these challenges, this paper presents “ShowerFlex: A friction and spring-balanced continuum hose with a showerhead at its distal end”. With ShowerFlex the showerhead can be positioned at any location and orientation within the bath space, and the hose holds its configuration without being continuously held or requiring external power supply. This allows for hands-free operation of the shower. Furthermore, our design allows the showerhead to be repositioned with little force, making critical areas such as the back and legs accessible to the shower. We refer to this hybrid strategy as pseudo-static balancing, a concept formally defined in Section 2.

Showerflex routes the hose through a Loc-Line® assembly, which is a series of hollow, friction‑locked ball‑and‑socket segments. One end of the Loc-Line connects to the water inlet, and the other attaches to a standard showerhead. A self‑locking retractable reel, anchored to a fixed support and connected to the showerhead, provides additional stability. The reel locks when stationary but allows smooth, user‑guided motion when slightly tensioned, while its internal springs help counteract the weight of the showerhead. Together, the Loc-Line and reel demonstrate the advantages of conventional static balancing, i.e. equilibrium at all configurations with minimal repositioning force.


The combination of friction locking and spring-loaded balancing is studied in this paper as a strategy to obtain the benefits of static balancing without actually requiring the forces to balance. To provide a theoretical basis for this behavior, we established a recursive forward kinematics model alongside a detailed static equilibrium analysis. These mathematical foundations enable numerical simulations, specifically using the Monte Carlo method to quantify the reachable workspace and verify the system's stability for normal bathing tasks. Finally, we demonstrate the robustness by physical experimental validation, confirming the device's intrinsic static stability and validating that the actuation force is well within the ergonomic capabilities of older adults.
\section{RELATED WORK}
Static balancing refers to maintaining a mechanical system in static equilibrium for all possible configurations by ensuring that all forces are balanced. A comprehensive theory for statically balancing linkages using zero‑length springs to exactly compensate end masses was presented by \cite{herder2001energy}. Static balancing offers two key practical design advantages: (i) stability, achieved through a constant potential energy leading to equilibrium in every configuration, and (ii) low actuation force, since only a small force is required to move the system. In practice, however, ideal balance, i.e. complete equilibrium combined with zero actuation force cannot be fully realized due to friction in the joints \cite{Kuo2021}. While joint friction contributes to stability by resisting unintended motion, it also increases the threshold actuation force required to move the mechanism \cite{Ramu2025}.

Static balancing has been applied to reduce the effective actuation stiffness of compliant mechanisms by incorporating members that exert forces in opposing directions \cite{Kuppens2021}. By combining negative‑stiffness elements with constant‑force elements, designers can achieve a quasi‑static balance over a defined operating range of the compliant mechanism \cite{10.1115/1.4002247}. It is also noteworthy that compliant mechanisms inherently lack friction, as their motion arises from elastic deformation rather than discrete joints.

Continuum links such as a shower hose possess theoretically infinite degrees of freedom, making true static balancing practically impossible. Loc‑lines, which are friction‑locked hollow ball‑and‑socket joints, maintain equilibrium through friction generated at the press‑fit interfaces. Related design embodiments, often referred to as goosenecks \cite{10.1093/asjof/ojz016}, use a hose tightly wrapped with a helical spring. Both of these commercially available mechanisms are highly effective in "set-and-forget" applications—such as directing machine tool coolant or positioning medical monitors—where frequent adjustment is unnecessary. In these cases, the design goal is to prioritize equilibrium and rigid holding capacity over minimizing actuation force.

However, when adapted for heavy loads with long moment arms, such as a showerhead extended beyond two feet, a fundamental mechanical limitation emerges: increasing the joint friction extends the holding range, but it proportionally increases the force required to reposition the device. To the best of our knowledge, there is no passive continuum or articulated mechanism reported in the literature capable of simultaneously achieving a stable cantilever hold exceeding about 4 ft while maintaining an average actuation force below 3 N.

For a shower hose supporting a showerhead at its end, both stability through equilibrium and low actuation force are essential. However, friction‑locked ball‑and‑socket joints inherently involve a trade‑off. For instance, a 1.6 kg showerhead must remain in stable equilibrium without collapsing, even when positioned as far as 4 ft from the inlet. Achieving this requires substantial frictional resistance. The closest flexible shower using this concept could not hold its length beyond two feet \cite{luettgen2003flexible}. Thus design also suffered from excessive force to reposition, but also suffered from noise. However, the user—often an older adult with limited upper‑body strength—may only be able to apply a modest actuation force to reposition the showerhead. 

Consequently, the competing requirements of stable equilibrium and minimal actuation force highlight a critical research gap. Existing accessible showering solutions fundamentally fall short because they either (a) require continuous user grip, which increases fall risks, (b) rely on external power and mechatronics, which introduce waterproofing and cost barriers, or (c) utilize purely passive friction joints that cannot hold extended configurations under typical showerhead loads without requiring excessive manual force. ShowerFlex directly addresses this gap through our proposed pseudo-static balancing strategy. Unlike traditional static balancing which mathematically zeroes out potential energy differences, our approach uses a spring-loaded retractor to bear the bulk of the gravitational moment, while relying on intrinsic joint friction to counteract only the residual unbalanced torque. This synergistic design resolves the long-standing trade-off, achieving both robust hands-free positioning and an ergonomically low repositioning force suitable for older adults.

\section{Design and Mode of Operation}

In this section, we describe the detailed mechanical design and the working principle of the proposed device. Showerflex consists of two components that work synergistically to provide stable equilibrium and reduce repositioning force. The frist component is a loc-line, which is a friction-locked ball and socket joint series, and the second is a retractable spring loaded reel. In this section, we explain the mechanics of these two components. Finally, we detailed the experimental procedures about how to get the friction parameters of each joint.

\begin{figure*}[t]
    \centering
    \begin{subfigure}{0.33\linewidth}
    \centering
    \includegraphics[height=5cm]{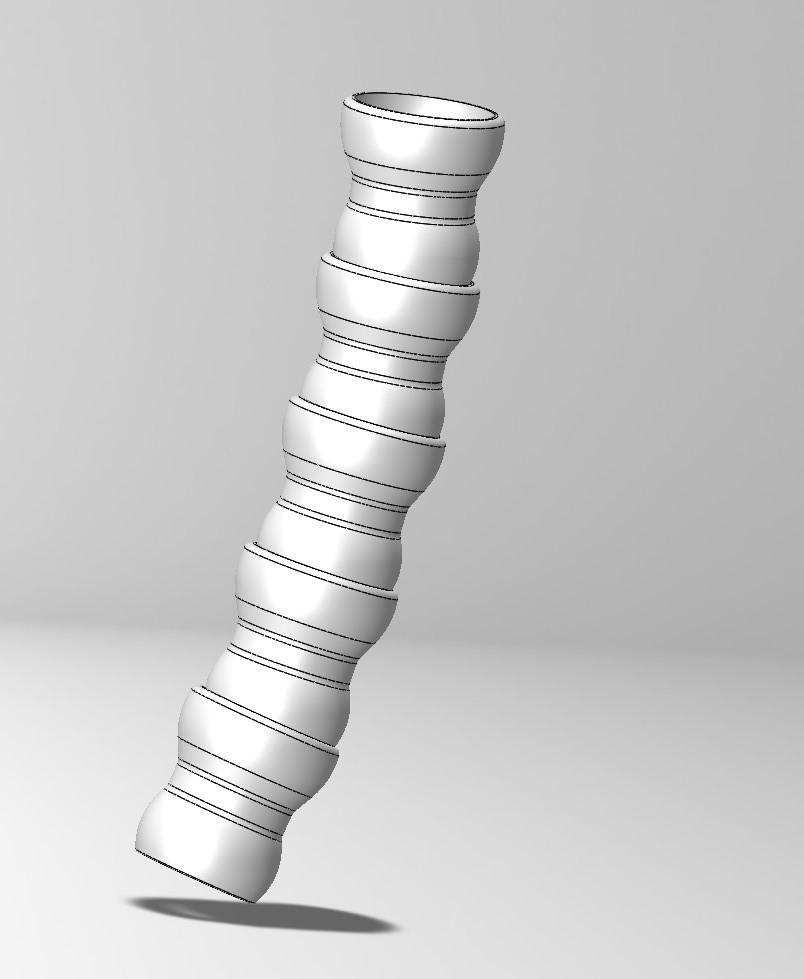}
    \caption{Multi-segment modular continuum assembly composed of ball-and-socket joints.}
    \label{fig:vertebra_unit}
    \end{subfigure}
    \hfill
    \begin{subfigure}{0.33\linewidth}
    \centering
    \includegraphics[height=5cm]{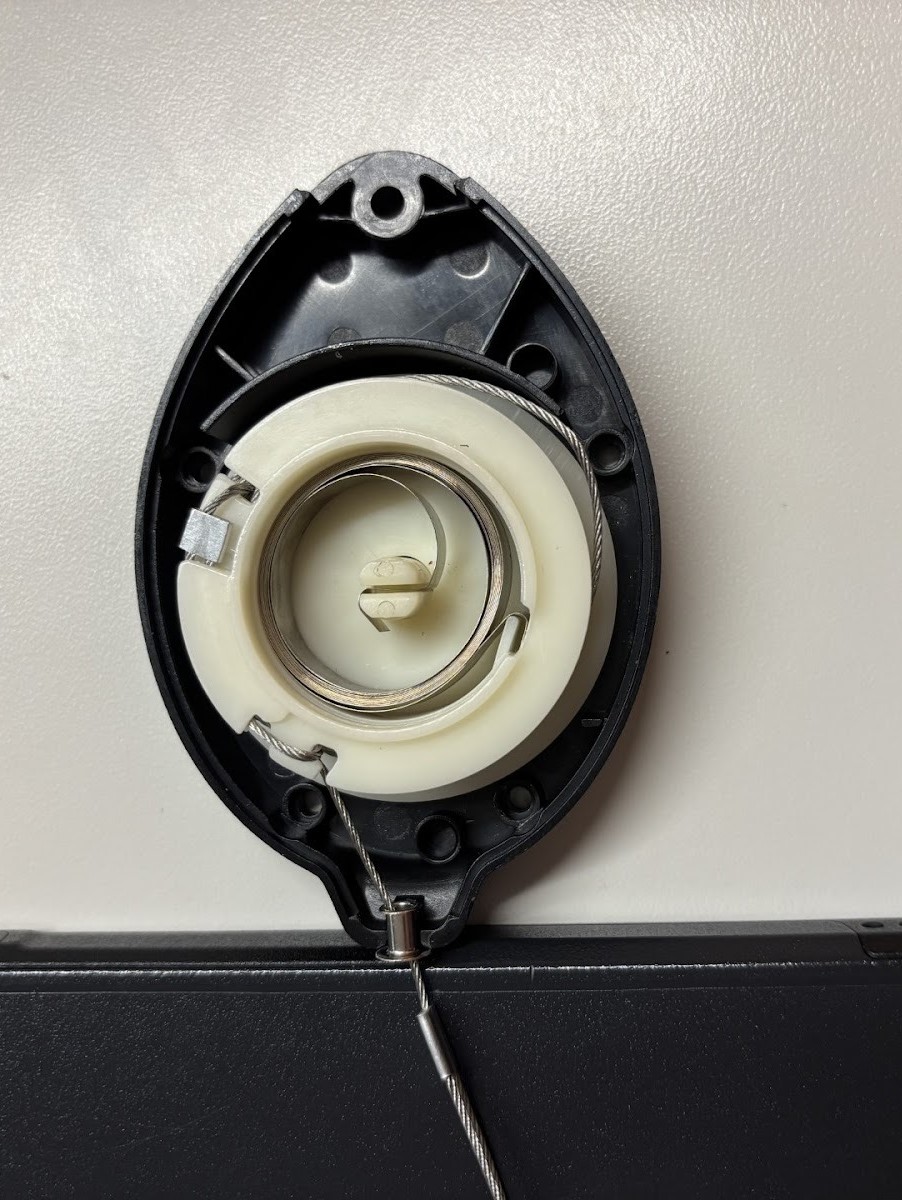}
    \caption{Internal mechanism of the retractable spring loaded reel.}
    \label{fig:Rectractor}
    \end{subfigure}
    \hfill
    \begin{subfigure}{0.33\linewidth}
    \centering
    \includegraphics[height=5cm]{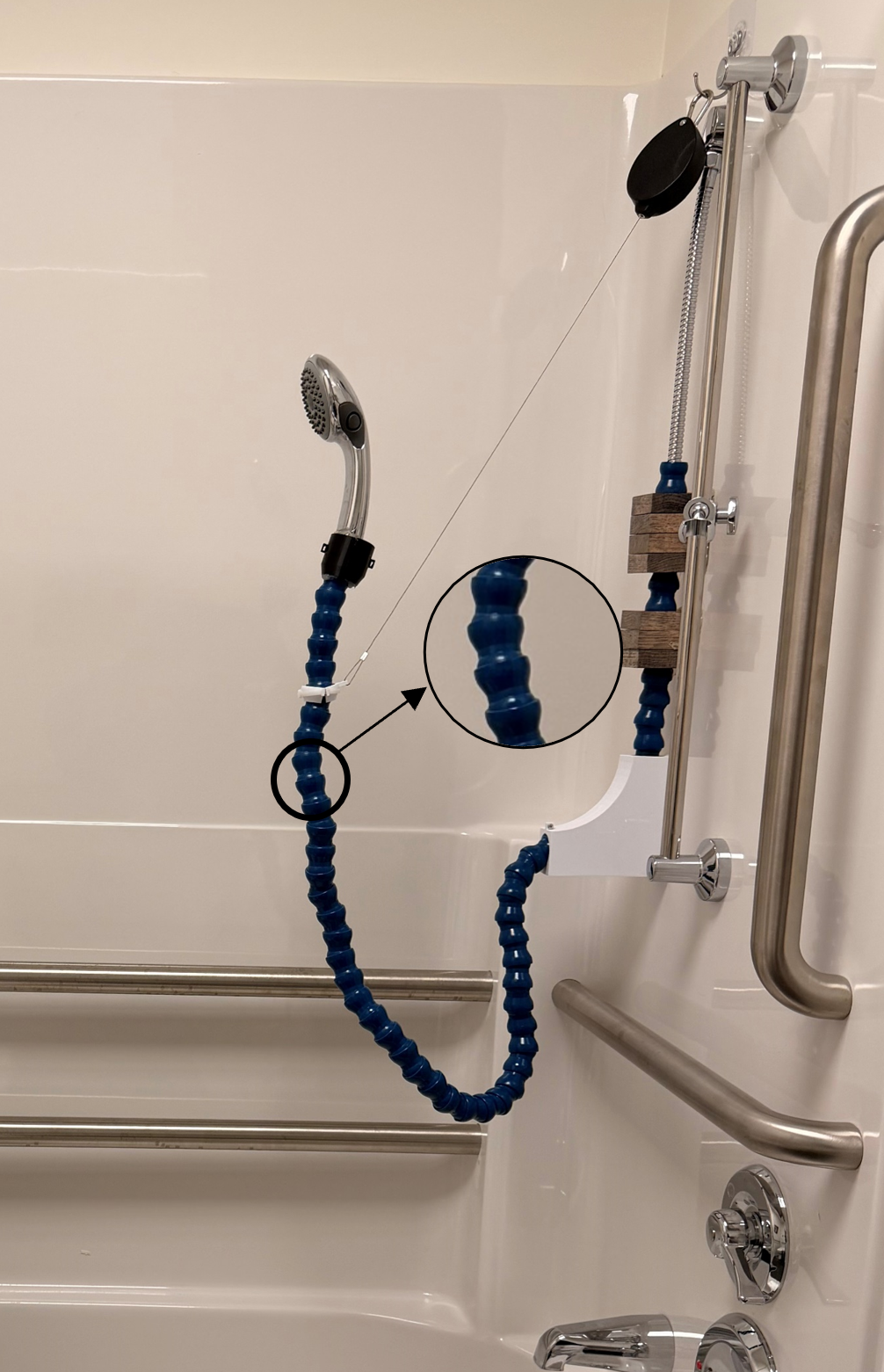}
    \caption{Installed the prototype in a bathtub environment.}
    \label{fig:overview}
    \end{subfigure}
    \caption{Physical components and prototype assembly of the statically balanced continuum mechanism.}
    \label{fig:vertebra_structure}
\end{figure*}

\subsection{LocLine Ball and Socket Joint Assembly}
The structural framework of the hose is synthesized from a series of identical hollow ball-and-socket joints, known as a Loc-Line. Each Loc-Line building block is made of high-density polymer with two hemispherical plastic surfaces with different diameters. The smaller hemispherical surface of one building block is press fit inside the larger hemispherical surface of another building block thus creating sufficient friction between the two. Multiple units are connected to form a kinematic sequence with dual degrees-of-freedom (DOF): \textbf{Yaw} ($Z$-axis) and \textbf{Pitch} ($Y$-axis) as shown in Fig.~\ref{fig:vertebra_unit}.


While each individual segment is rigid and non-deformable, the collective assembly approximates a continuum arm. In our design, we use 51 building block segments to make up the arm. This discretized architecture provides the necessary multi-DOF flexibility required to navigate and conform to the complex, non-linear contours of the human back and limbs during bathing tasks. While the friction between the building blocks can keep the entire continuum hose in equilibrium, adding a heavy showerhead at the distal end will cause the moment in some joints to overcome the friction. This will lead to the collapse of the hose. Excessive friction also leads to noise while repositioning, which needs to be avoided.

\subsection{Spring-Loaded Retractor and Self-Locking Mechanism.}
To provide dynamic tensioning and state-dependent position holding, the system incorporates a mechanical spring-loaded retractor equipped with a velocity-sensitive self-locking mechanism . The core architecture of this module consists of a pre-tensioned torsion spring coupled to a rotational spool, integrated with a centrifugal ratchet-and-pawl assembly as shown in Fig.\ref{fig:Rectractor}.

The operational state of the retractor is entirely governed by the input applied to the cable. During rapid linear extraction, the high angular velocity of the spool generates sufficient centrifugal force to keep the internal pawl disengaged from the ratchet gear, enabling smooth, continuous unspooling with minimal resistance. Conversely, when the extraction speed decelerates or halts, the centrifugal force reduces allowing a built-in biasing mechanism to drive the pawl into the nearest ratchet tooth, immediately locking the spool. In this engaged state, the mechanism physically blocks the restoring torque of the torsion spring, maintaining a fixed cable length. To disengage the lock and initiate retraction, a momentary distal displacement is applied. This micro-movement lifts the pawl out of the locking groove, allowing the torsion spring to freely rewind the cable into the spool housing. The purely mechanical logic in the reel ensures robust, power-free state transitions between dynamic extension and static locking. The reel thus serves two purposes: aiding equilibrium when the pawl is locked, and offsetting gravity when in motion thus reducing the repositioning force.

The combination of the loc-line and the spring-loaded retractable reel, integrated into the complete physical prototype shown in Fig.~\ref{fig:overview}, provide static stability to the showerhead at multiple positions at the \textbf{release mode}, while yet reducing the force needed to reposition in the \textbf{manipulation mode}. During the release mode, the user applies no force on the showerhead. Here, the combination of the friction in the locline joints and the friction between the pawl and ratchet gear offsets the weight of the showerhead. This effectively ``self-locks'' the mechanism in its new position, ensuring it remains stationary without electronic brakes or manual locking screws. \textbf{During manipulation mode}, the external force applied by the user dislodges the pawl from the ratchet gear, thus engaging the spring force from the reel. The user applies sufficient force to overcome dynamic friction between the ball and socket joints of the loc-line to move it to a particular position. When the movement of the showerhead stops, the pawl engages with the ratchet, and the whole device is in equilibrium.

\section{Analysis Methods}
In this section, we establish the mathematical foundations that can be used to model the Loc-Line mechanism. The aim of the model is to establish the (a) static stability of showerhead at different useful showering positions, and (b) evaluate forces exerted to reposition the showerhead. Our analysis methods will deal with just the first aspect, while the second will be validated using experiments. 
\begin{figure}[b]
    \centering
    \includegraphics[height=5cm]{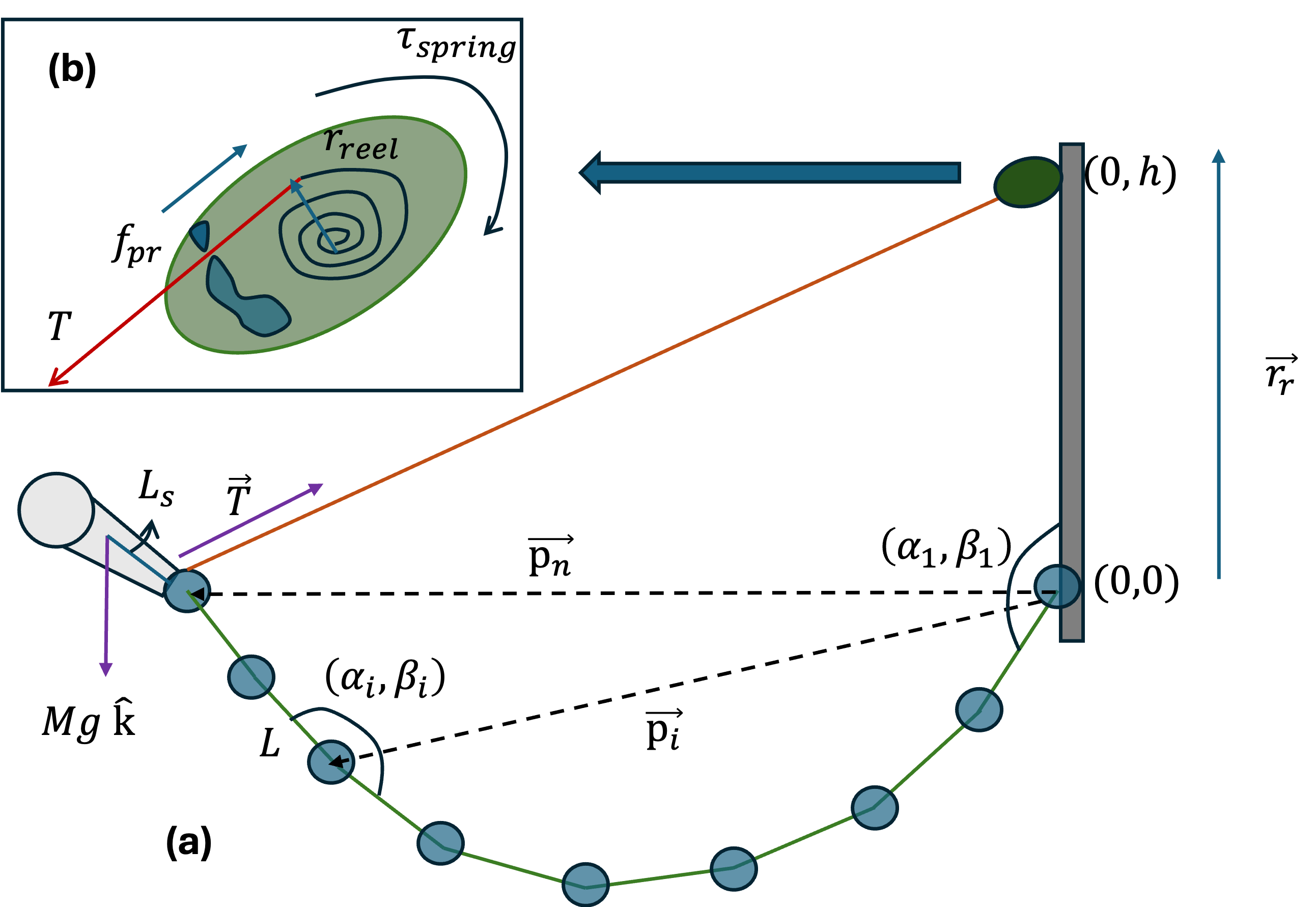}
    \caption{(a) Forward kinematic parameters for a loc-line, with (b) the forces in a spring loaded reel}
    \label{fig:Kinematics}
\end{figure}
\subsection{Kinematic modeling of the continuum loc-line}
The geometric configuration of the Loc-Line continuum is defined by a discrete chain of $n$ identical modular segments as shown in Figure \ref{fig:Kinematics}a. To precisely map the spatial posture of the mechanism, we employ a recursive coordinate transformation framework. Each joint possesses two rotational degrees of freedom (DOF): a yaw angle $\alpha_i$ and a pitch angle $\beta_i$, corresponding to rotations about the local $y$ and $z$ axes, respectively.

The orientation of the $i$-th segment relative to its predecessor is determined by the composition of two elementary rotation matrices. Based on the mechanical constraints of the ball-and-socket joints, the yaw rotation matrix $\bf{R}_{y,i}$ and the pitch rotation matrix $\bf{R}_{z,i}$ are defined as:
\begin{align}
\mathbf{R}_{y,i}(\alpha_i) = \begin{bmatrix} 
\cos\alpha_i & 0 & \sin\alpha_i \\ 
0 & 1 & 0 \\ 
-\sin\alpha_i & 0 & \cos\alpha_i 
\end{bmatrix} \\
\mathbf{R}_{z,i}(\beta_i) = \begin{bmatrix} 
\cos\beta_i & -\sin\beta_i & 0 \\ 
\sin\beta_i & \cos\beta_i & 0 \\ 
0 & 0 & 1 
\end{bmatrix}
\end{align}

The local transformation of a single segment is thus given by the product of these elementary rotations:
\begin{equation}
\mathbf{R}_{seg,i} = \mathbf{R}_{y,i}(\alpha_i) \cdot \mathbf{R}_{z,i}(\beta_i)
\end{equation}

Given that the mechanism is a serial kinematic chain, the absolute orientation of any segment $i$ within the global reference frame is determined by the cumulative product of all preceding joint rotations. We define the accumulated rotation matrix $\mathbf{R}_{acc,i}$ through the following recursive relation:
\begin{equation}
\mathbf{R}_{acc,i} = 
\begin{cases} 
\mathbf{I}_{3 \times 3}, & \text{if } i = 0 \\
\mathbf{R}_{acc,i-1} \cdot \mathbf{R}_{seg,i}, & \text{if } i = 1, \dots, n 
\end{cases}
\end{equation}

where $\mathbf{I}_{3 \times 3}$ represents the identity matrix at the fixed base $\mathbf{p}_0 = [x_0, y_0, z_0]^T$. This serves as the initial orientation of the system before any joint deformation is applied. The spatial position of each subsequent node $\mathbf{p}_{i+1}$ is propagated along the transformed local $x$-axis of the current segment. According to the principle of vector summation, the position of the $(i+1)$-th node is derived as:
\begin{equation}
\vec{p}_{i+1} = \vec{p}_i + L \cdot \left( \mathbf{R}_{acc,i} \cdot \mathbf{e}_1 \right)
\end{equation}

where $L$ is the invariant length of each module and $\mathbf{e}_1 = [1, 0, 0]^T$ is the unit vector along the longitudinal axis of the segment. By iterating this process for $i = 1, \dots, n$, the coordinates of the entire spine and the end-effector pose are uniquely determined by the joint angle set $\{\alpha, \beta\}_{1:n}$.

\subsection{Static equilibrium}
The mechanism is in static equilibrium when the moment acting at each joint of the loc-line due to the weight of the showerhead is balanced by the friction in the ball and socket and the tension in the retractable reel. Let the center of gravity of the showerhead be located at $\vec{p}_s$ from the origin, as shown in Fig. \ref{fig:Kinematics}a. Let the retractable reel be vertically above the origin at a distance $h$ given by the position vector $\vec{p}_r$. The net external moment acting at joint $i$ is given by 
\begin{equation}
    \mathbf{M}_i=-M_s g\hat{\mathbf{e}}_3\times(\vec{p}_s-\vec{p}_i)+T\frac{\vec{p}_r-\vec{p}_n}{||\vec{p}_r-\vec{p}_n||}\times(\vec{p}_n-\vec{p}_i)
    \label{Eq:NetMoment}
\end{equation}
In the above equation, it is assumed that the cable of the retractable reel is fixed at the $n^{th}$ joint of the loc-line. For equilibrium, the moment $M_i$ must be balanced by the frictional torque in the ball and socket joint.
\begin{align}
&\forall i = 1\to n \nonumber\\
&\mathbf{M}_i-\tau_{fi}\frac{\mathbf{M}_i}{||\mathbf{M}_i||}=\mathbf{0}
\label{Eq:Equilibirum}\\
&s.t. \tau_{fi}\leq \tau_{fmax} \nonumber
\end{align}
where $\tau_{fi}$ is the frictional torque in each loc-line acting in the opposite direction of the net external moment. The static frictional torque has an upper limit $\tau_{fmax}$, which when exceeded, begins to move the mechanism. The equilibrium equation can be restated to show the existence of the reel tension $T$ for a given configuration (or joint angles) of the locline such that the equilibrium of Eq. \ref{Eq:NetMoment} and \ref{Eq:Equilibirum} are satisfied at a specific joint $k$ in the limiting friction case where $\tau_{fk}=\tau_{max}$. 

The reel tension $T$ itself is internally balanced by the spring restoring force and the friction force of the pawl and the ratchet as shown in Fig. \ref{fig:Kinematics}b. It is clear from these equations that the equilibrium is dependent on the maximum frictional torque that each joint can exert to resist rotation. This is characterized by the experiment below.

\subsection{Joint Friction Characterization}
To accurately model the mechanism's holding capacity, we conducted an experiment to quantify the static friction torque of a ball-and-socket joint. To minimize measurement errors caused by the short length of a single segment, we utilized a multi-segment assembly consisting of four nodes. We mechanically locked the proximal nodes to create a rigid extension, allowing rotation only happened at the most distal joint. This setup significantly increased the effective moment arm $L$, ensuring higher measurement precision.

When the load added is less than the static friction limit of the joint, there was no joint movement. However, once the static limit exceeds, the joint collapses. 

\begin{figure}[h]
    \centering
    \includegraphics[height=6cm]{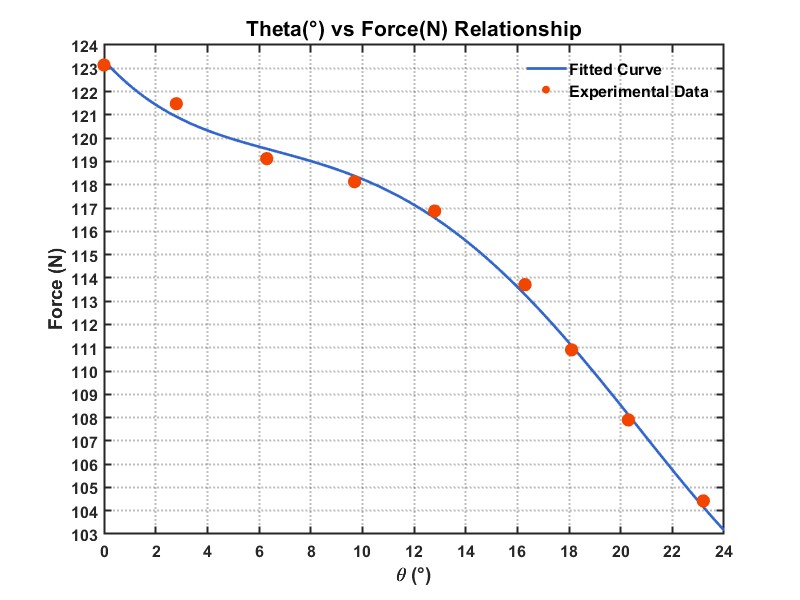}
    \caption{Experimentally characterized relationship between static friction force and joint bending angle.}
    \label{fig:Friction}
\end{figure}

Due to this discrete behavior, we adopted a static threshold testing method. We manually preset the active joint to specific angles (e.g., $0^{\circ}, 6.3^{\circ}, 12.8^{\circ}, \dots, 23.2^{\circ}$). For each angle, we hung varying weights on the end-effector to find the critical balanced mass that kept the system at the boundary between static equilibrium and motion. Using the torque equilibrium formula:
\begin{equation}
m \cdot g \cdot L \cdot \cos(\theta) = f \cdot r
\end{equation}

where $m$ is the balanced mass, $L$ is the moment arm, and $r$ is the joint radius, we calculated the specific friction torque $f \cdot r$. The results indicate that friction is highest at the neutral position and decreases as the bending angle increases. We fitted a polynomial curve to this data to represent the joint friction in our simulation model as shown in Fig.~\ref{fig:Friction}.

\begin{figure*}[b]
    \centering
    \begin{subfigure}{0.4\textwidth}
        \centering
        \includegraphics[width=\textwidth]{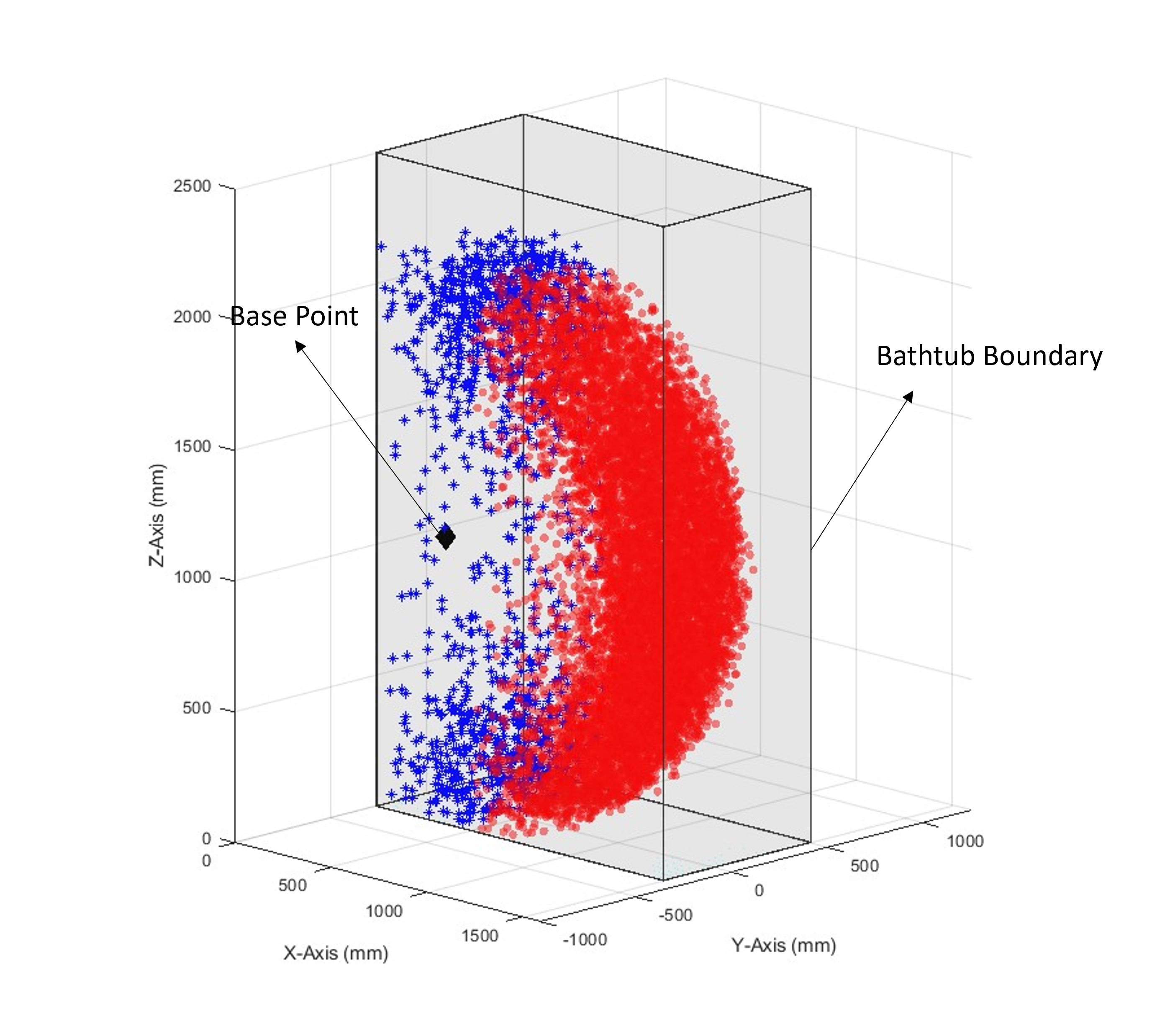}
        \caption{The Workspace without spring-loaded reel.}
        \label{fig:Without}
    \end{subfigure}
    \hspace{0.5cm} 
    \begin{subfigure}{0.4\textwidth}
        \centering
        \includegraphics[width=\textwidth]{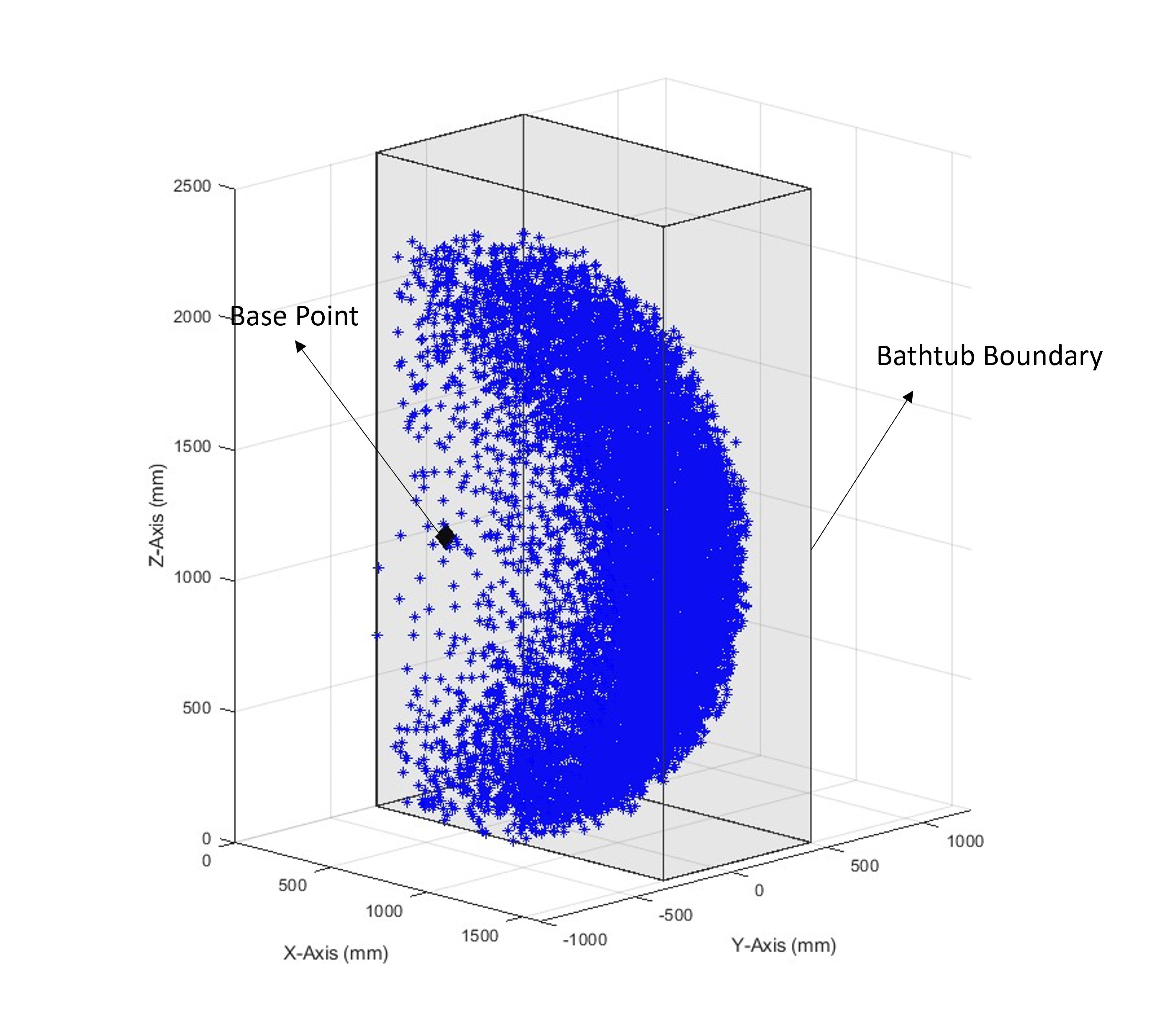}
        \caption{The Workspace under spring-loaded reel.}
        \label{fig:with}
    \end{subfigure}  
    \caption{The workspace comparison between the device with reel and without reel.}
    \label{fig:comparison}
\end{figure*}
\section{RESULTS}
In this section, we validate the stability and ease of repositioning the showerhead using the ShowerFlex mechanism. The validation can be divided into two parts: numerical simulation and physical experimentation. First, we use the forward kinematic model to analyze the workspace and demonstrate stability through equilibrium in specific bathing configurations. Second, we test the physical prototype to verify the device's stability at these configurations. Finally, we experimentally measure actuation force required for repositioning between two configurations. These results show the feasibility and effectiveness of the system.

\subsection{Kinematic Workspace Reachability Analysis}

To estimate and evaluate the workspace at the typical bathroom environment, we conducted a numerical analysis by Monte Carlo method. This method generates a large number of random joint points and then checks for the equilibrium condition of Eq. \ref{Eq:Equilibirum} . We collected all the stable end-effector points to map a point clouds in Figure \ref{fig:comparison} to show the workspace reachability.

The simulation parameters were set to match the physical prototype described in Section 4. The mechanism consists of $N=51$ modular segments, each with a unit length of $L=25$ mm. The base of the mechanism is fixed at the wall position $\mathbf{P}_{base} = [100, 0, 1000]^T$ mm. The joint angle limits for every ball and socket joint were set to $|\theta| \le 23.2^{\circ}$.

We generated $N_{samples}=20,000$ random samples to ensure sufficient density for the workspace analysis. For each sample, the yaw ($\alpha$) and pitch ($\beta$) angles of 51 joints were given to random values within the joint range using a uniform distribution. The end-effector position for each configuration was calculated by the Forward Kinematics formula which was derived in Eqn. (5). We collected all the end-effector position to map a point cloud figure.

To determine if a configuration is valid, we applied two filters based on the environment and the load-balancing logic:
\begin{enumerate}
    \item \textbf{Geometric Boundaries:} The end-effector must stay within the bathtub boundaries which are defined by $x \in [0, 1500]$ mm, $y \in [-260, 510]$ mm, and $z > 0$ mm.
    \item \textbf{Static Stability:} We checked the stability of each point. Points were classified as ‘Stable' (colored by Blue) if equilibrium equations are satisfied. Configurations where the gravitational moment arm exceeded the system's holding capacity were classified as ‘Unstable' (colored by Red).
\end{enumerate}

\begin{figure*}[t]
    \centering
    \begin{subfigure}{0.23\textwidth}
        \centering
        \includegraphics[width=\textwidth]{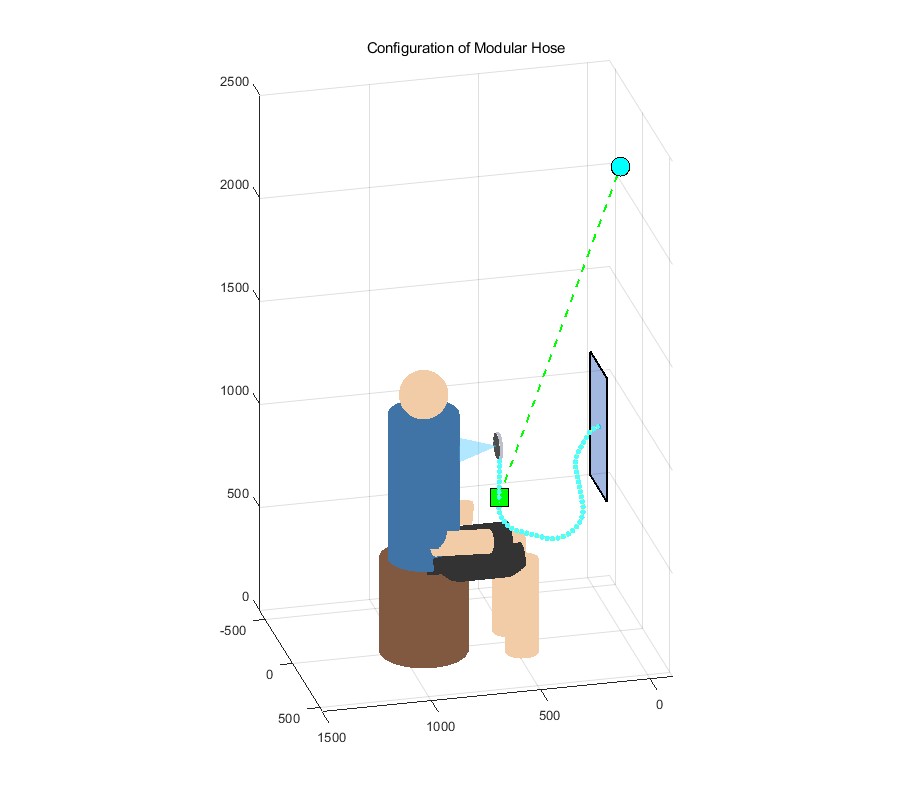}
        \caption{Simulated Configuration for Front Bathing.}
        \label{fig:1}
    \end{subfigure}
    \hspace{0cm}
    \begin{subfigure}{0.23\textwidth}
        \centering
        \includegraphics[width=\textwidth]{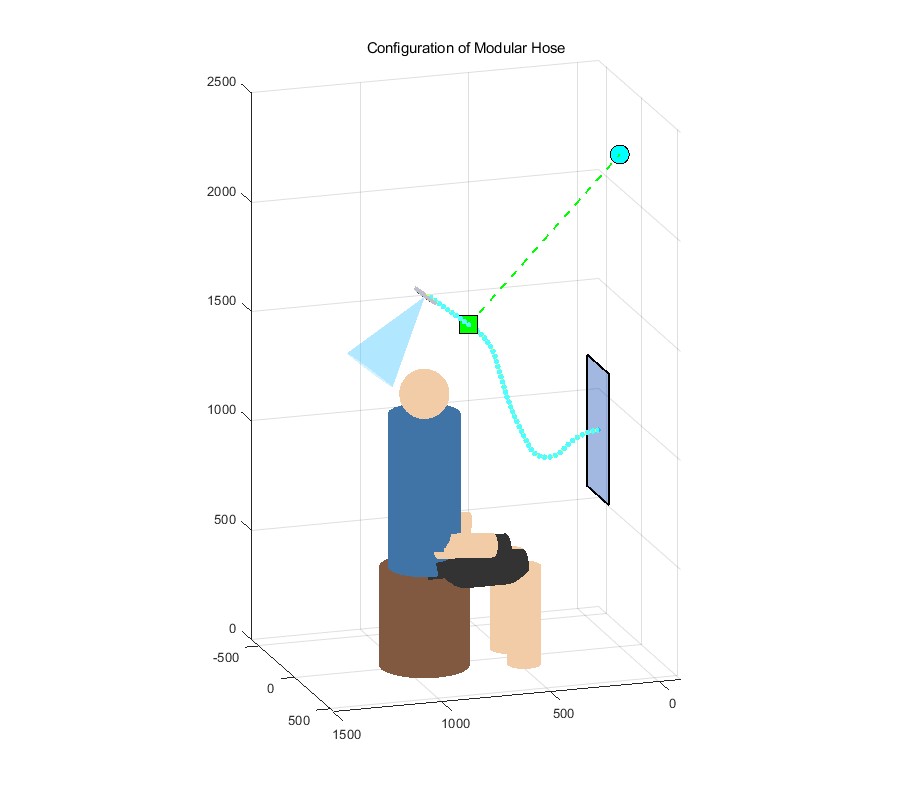}
        \caption{Simulated Configuration for Head Bathing.}
        \label{fig:2}
    \end{subfigure}
    \hspace{0cm}
    \begin{subfigure}{0.23\textwidth}
        \centering
        \includegraphics[width=\textwidth]{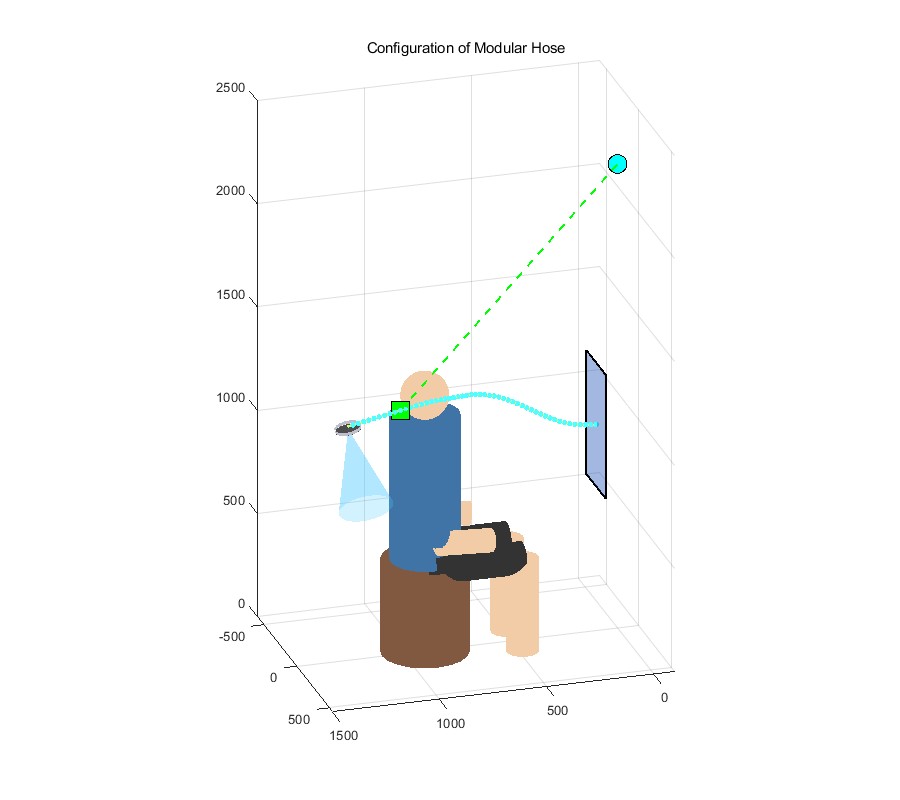}
        \caption{Simulated Configuration for Back Bathing.}
        \label{fig:3}
    \end{subfigure}
    \hspace{0cm}
    \begin{subfigure}{0.23\textwidth}
        \centering
        \includegraphics[width=\textwidth]{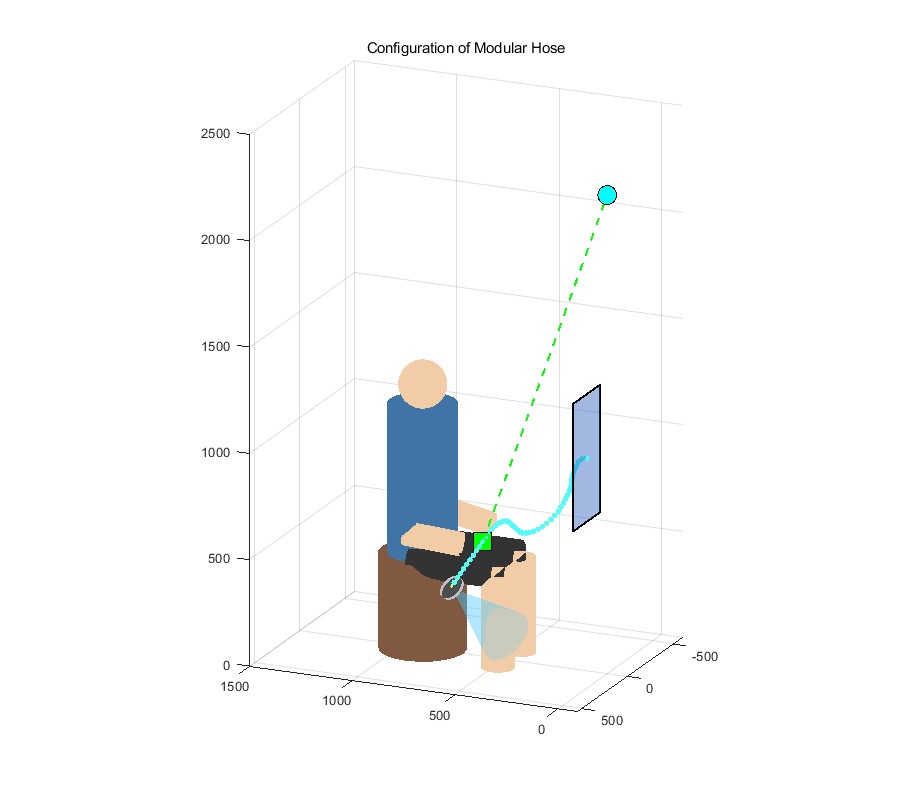}
        \caption{Simulated Configuration for Leg Bathing.}
        \label{fig:4}
    \end{subfigure}
   
    \caption{Simulated bathing configurations for common activities of daily living (ADLs).}
    \label{fig:4}
\end{figure*}

\begin{figure*}[t]
    \centering
    \begin{subfigure}{0.2\textwidth}
        \centering
        \includegraphics[width=\textwidth]{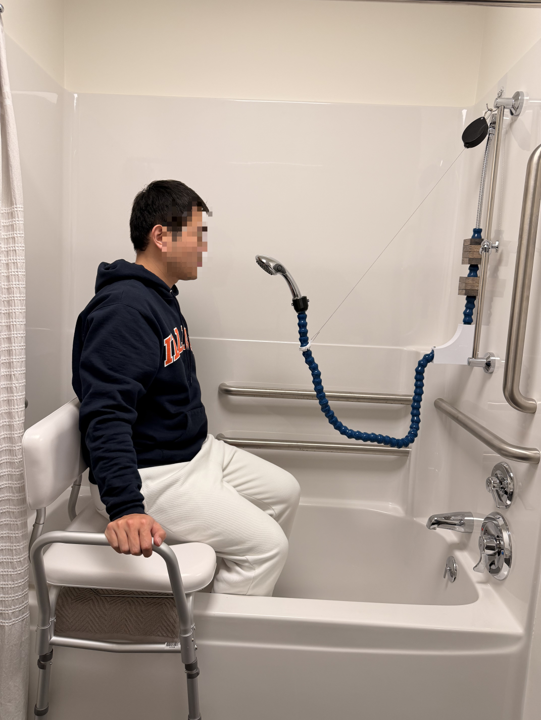}
        \caption{Experimental Configuration for Front Bathing.}
        \label{fig:1}
    \end{subfigure}
    \hspace{1cm}
        \begin{subfigure}{0.2\textwidth}
        \centering
        \includegraphics[width=\textwidth]{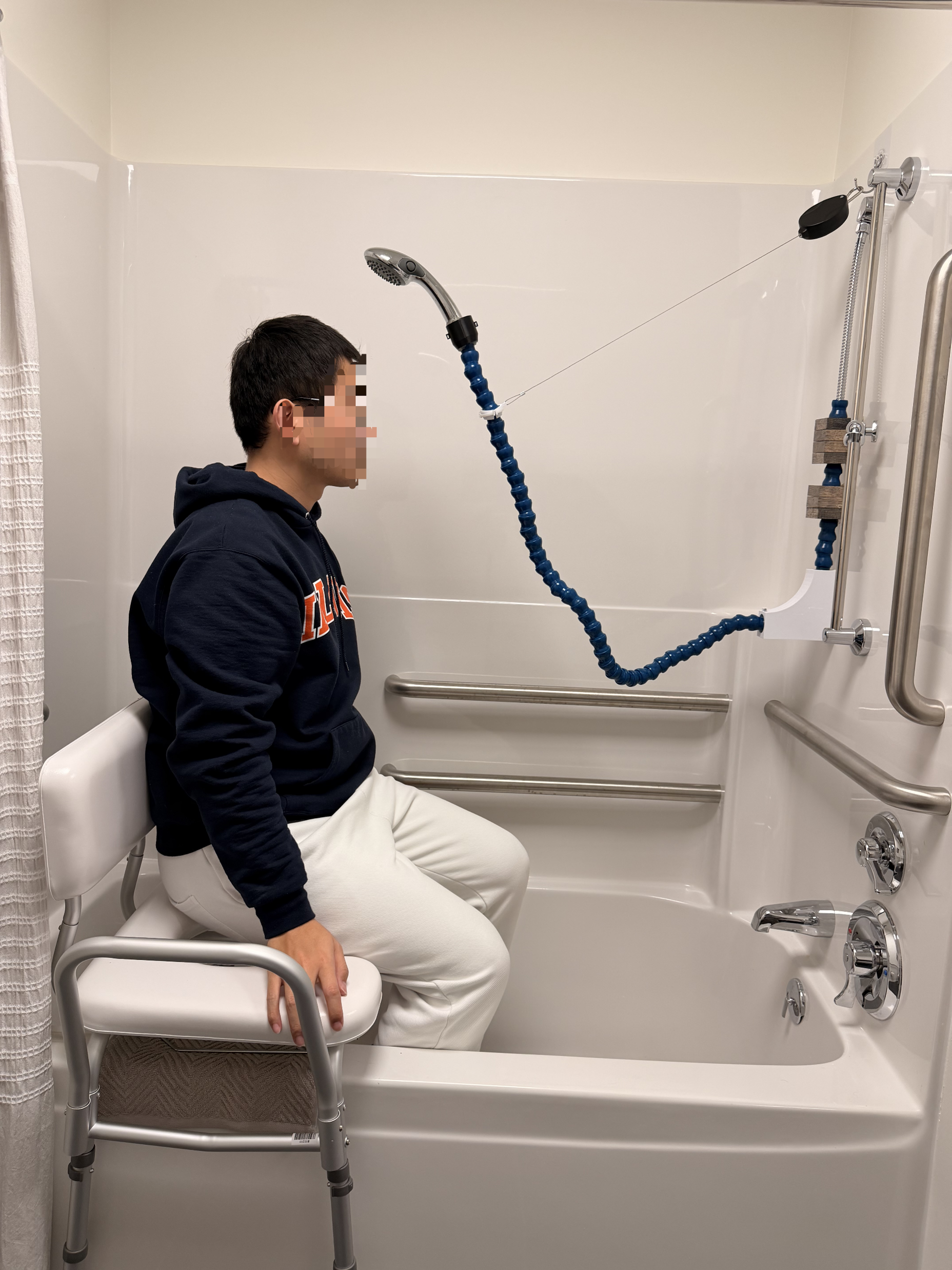}
        \caption{Experimental Configuration for Head Bathing.}
        \label{fig:1}
    \end{subfigure}
    \hspace{1cm}
    \begin{subfigure}{0.2\textwidth}
        \centering
        \includegraphics[width=\textwidth]{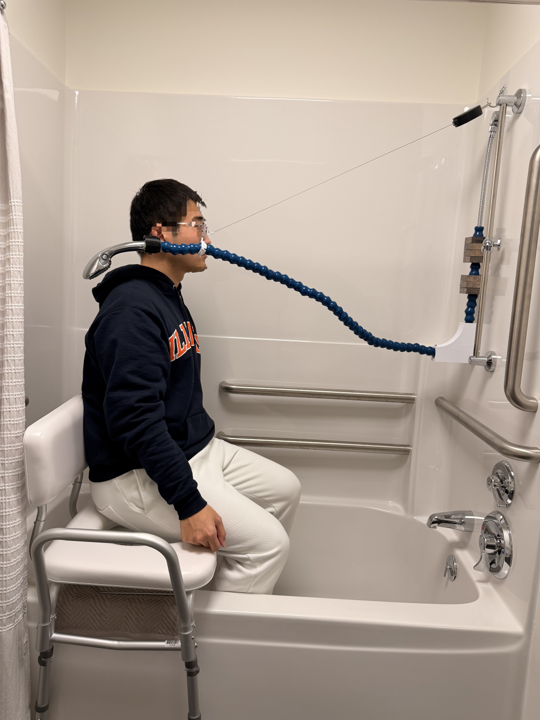}
        \caption{Experimental Configuration for Back Bathing.}
        \label{fig:2}
    \end{subfigure}
    \hspace{1cm}
    \begin{subfigure}{0.2\textwidth}
        \centering
        \includegraphics[width=\textwidth]{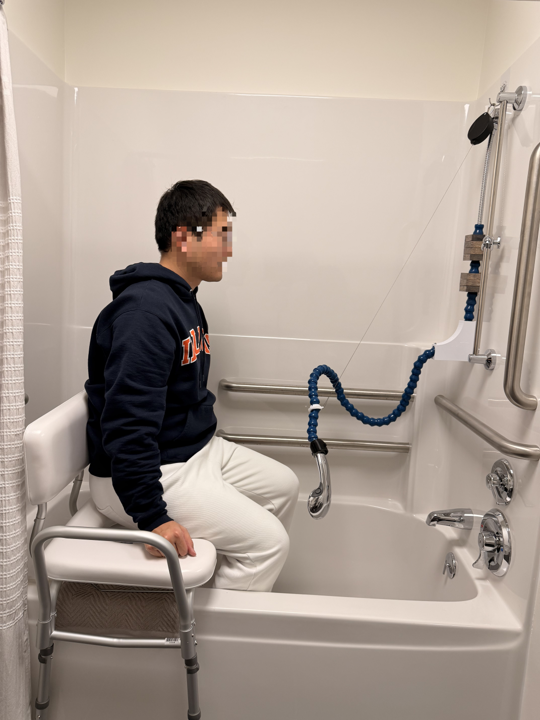}
        \caption{Experimental Configuration for Leg Bathing.}
        \label{fig:3}
    \end{subfigure}
   
    \caption{Experimental bathing configurations for common activities of daily living (ADLs).}
    \label{fig:5}
\end{figure*}

To validate the effectiveness of the mechanical design, we compared the workspace reachability under two different conditions:

\begin{enumerate}
    \item \textbf{Scenario A (Without Reel):} The system only relies on the internal joint friction to resist gravity of end-effector. The workspace result was shown in Fig.~\ref{fig:Without}

    \item \textbf{Scenario B (With Reel):} The system utilizes the proposed spring-loaded reel and internal joint friction for gravity compensation. The workspace result was shown in Fig.~\ref{fig:with} 
\end{enumerate}

The generated workspace point clouds, as shown in Fig.~\ref{fig:comparison}, visually demonstrate this comparison. Quantitatively, out of the 20,000 generated Monte Carlo samples, 10,767 configurations (53.84\%) fell within the valid bathtub boundaries. Relying solely on joint friction (without the reel), the system maintained static stability in only 1,217 (11.3\%) of these configurations. In contrast, integrating the spring-loaded reel rendered all 10,767 (100\%) valid configurations completely stable. This represents an almost 9-fold quantitative expansion of the usable reachable workspace. This confirms that the spring-loaded reel is essential for covering the necessary height range for both front (head/chest washing) and back (back/shoulder washing) positions.
\begin{figure*}[t]
    \centering
    \begin{subfigure}{0.46\textwidth}
        \centering
        \includegraphics[width=6cm, angle=-90]{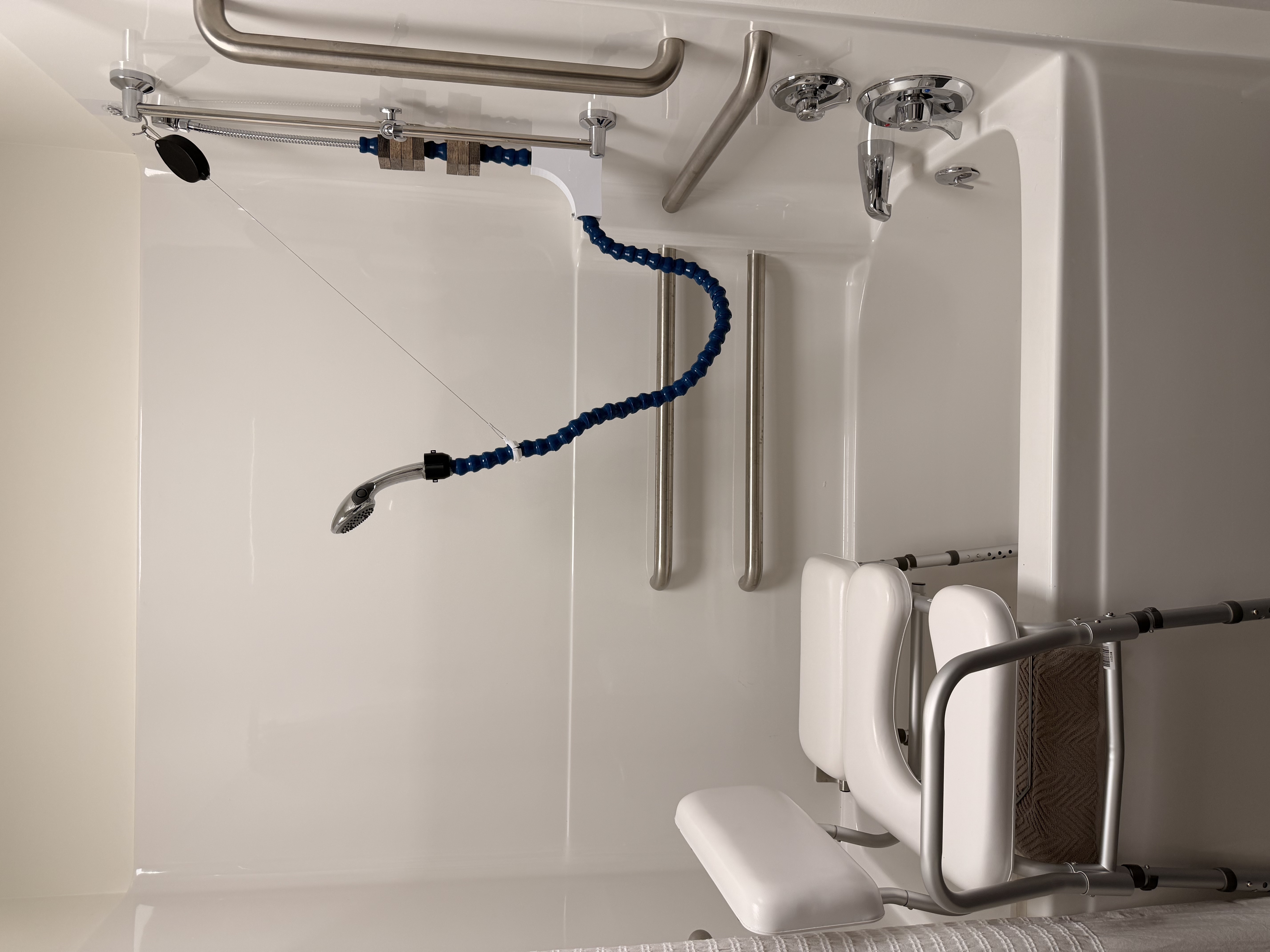}
        \caption{Starting configuration for force measurement.}
        \label{fig:position1}
    \end{subfigure}
    \hspace{0.1cm}
    \begin{subfigure}{0.46\textwidth}
        \centering
        \includegraphics[width=6cm, angle=-90]{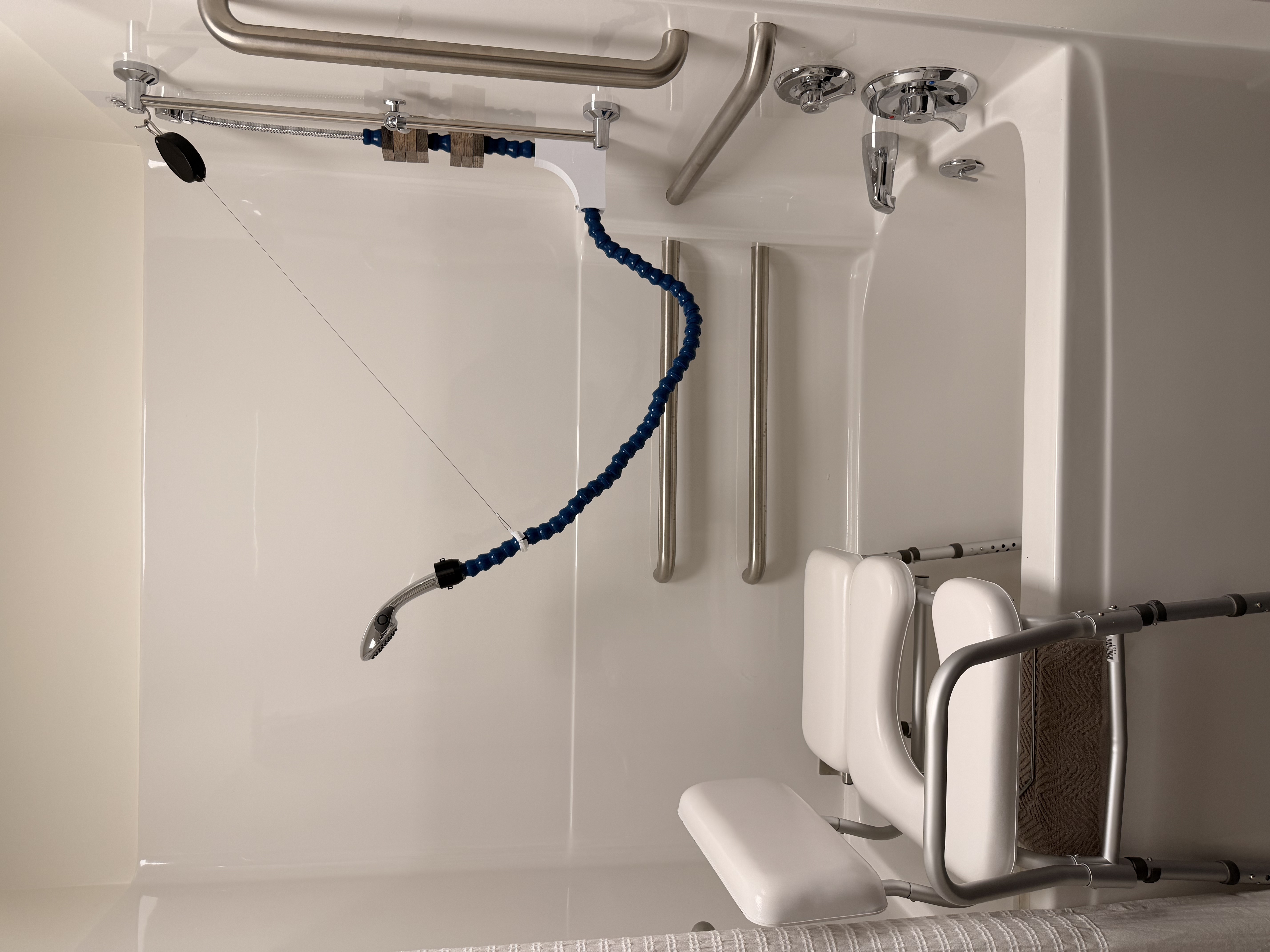}
        \caption{Ending configuration for force measurement.}
        \label{fig:position2}
    \end{subfigure}
    \caption{Starting and Ending Configurations for the Actuation Force Assessment}
    \label{fig:position}
\end{figure*}

\begin{figure*}[t]
    \centering
    \includegraphics[width=0.6\textwidth]{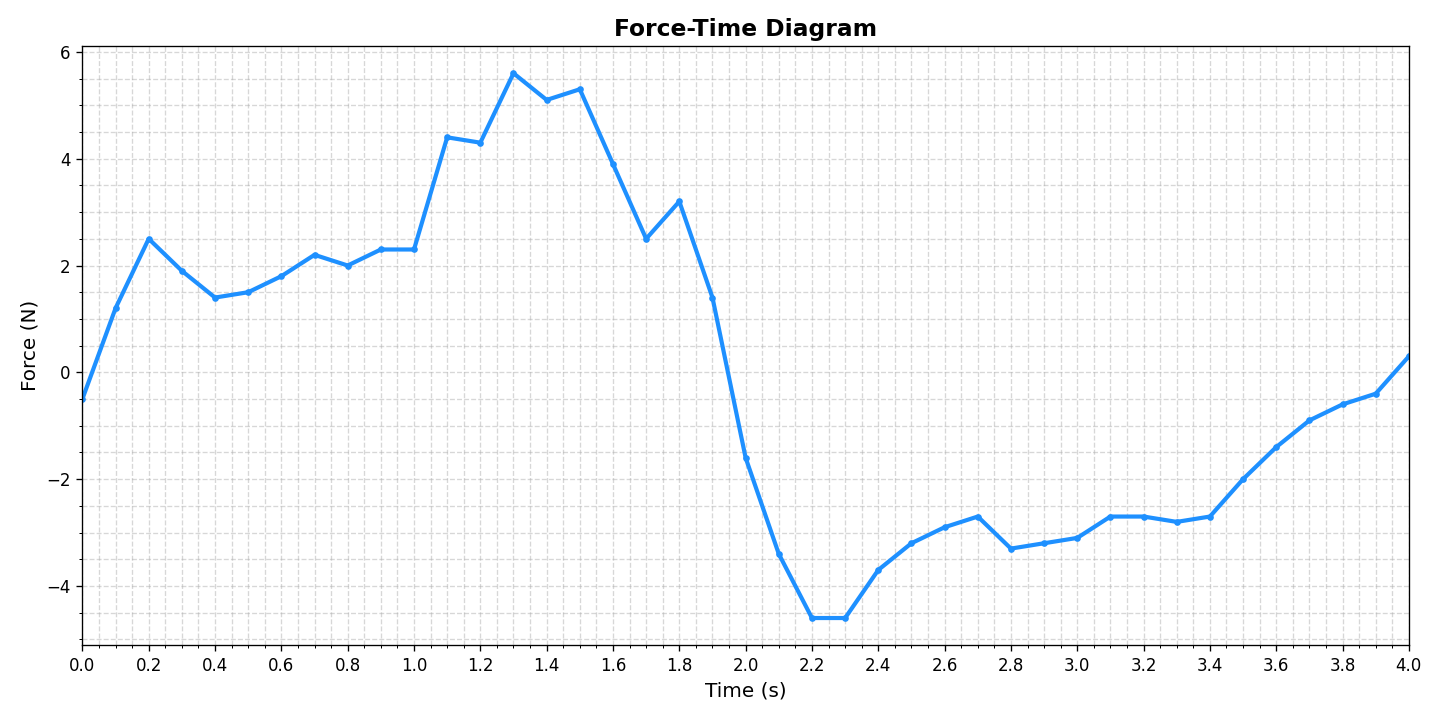}
    \caption{Force-time profile measured during the manual repositioning of the showerhead.}
    \label{fig:Force_data}
\end{figure*}

\subsection{Configuration Simulation for Daily Hygiene Tasks}

While the workspace analysis and comparison only show where the end-effector can reach, it is also very vital for us to figure out if the device will collide with the users or block their movements when they are sitting. That's why we need to use Forward Kinematics to simulate the complete spatial configuration for four daily hygiene tasks: front washing, head washing, back washing, and leg washing. 

\textbf{Simulation Setup and Results:}
By inputting specific yaw ($\alpha$) and pitch ($\beta$) angles, we generated the full posture of all 51 segments. The simulation figures show that the device can be used to bath user in a flexible way as shown in Fig.~\ref{fig:4}.
\begin{enumerate}
    \item \textbf{Front Bathing:} 
    The continuum mechanism arches upward and extends forward, smoothly positioning the showerhead to comfortably reach the user's anterior torso.

    \item \textbf{Head Bathing:} 
    The mechanism extends upward and overhead, precisely directing the water flow to facilitate hair and face washing without requiring the user to lean forward.

    \item \textbf{Back Bathing:} 
    This configuration highlights the advantage of the highly articulated continuum structure. While a traditional rigid straight arm would be physically obstructed by the user's body, our proposed mechanism safely curves around the torso to prevent collisions and reach the back seamlessly.

    \item \textbf{Leg Bathing:} 
    The mechanism smoothly bends downward, granting easy and targeted access to the lower extremities while the user remains safely and comfortably seated.
    
\end{enumerate}

\textbf{Conclusion:}
By simulating the full configurations, we verified that the device is not only capable of reaching target points, but also geometrically compatible with the human body. It can adapt to different configurations without causing any spatial interference when users are sitting.

\subsection{Experimental Validation: Stability and Operation Force}
To bridge the gap between simulation and reality, we conducted physical experiments using the prototype described in Section 3.2. The experiments focused on two critical aspects for elderly users: static stability and operation force.

\subsubsection{Static Stability Verification}
\hspace*{\parindent}
We positioned the device in three representative configurations corresponding to the daily hygiene tasks simulated in Section 5.2: Front, Head, Back, and Leg. A human subject sat in a standard bathtub environment to demonstrate these scenarios as shown in Fig.~\ref{fig:5}.

The mechanism successfully maintained all three postures without any active manual lock. In the \textbf{Front}, \textbf{Head}, and \textbf{Back} washing positions, the device remained stable against gravity, confirming that the spring-loaded reel effectively neutralized the weight. In the \textbf{Leg} washing position, the device held its downward extension without retracting. Throughout the test, no significant drift or collapse was observed, verifying the reliability of the “Push-and-Stay" logic in a real-world setting.

\subsubsection{Actuation Force Assessment}
\hspace*{\parindent}
To ensure the device is compliant with the physical capabilities of older adults, we measured the actuation force which can actuate or reshape the mechanism. Using a digital force gauge, we recorded all the data information during operation.

The measurement process involved pushing and pulling the end-effector, which was connected to the force gauge, from a starting configuration to an ending configuration, as depicted in Fig.~\ref{fig:position}. The corresponding force-time data recorded during this movement is shown in Fig.~\ref{fig:Force_data}. The results indicated that the average actuation force was approximately {2.6N}. This value is significantly lower than the maximum force capacity of an average older adult (typically around 30 N). The low actuation force confirms that the internal friction is well-balanced by the reel, ensuring that users can manipulate the device without fatigue.

\section{CONCLUSIONS}

This paper introduced Showerflex: a highly articulated continuum showerhead designed to assist older users with independent showering. Through the integration of a modular ball and socket loc-line and a spring-loaded retractable reel, we proposed a pseudo-static balancing strategy to achieve a “Push-and-Stay" functionality for the showerhead with no external power. This behavior was captured by a recursive forward kinematics model and static equilibrium analysis. The kinematic modeling allowed us to simulate the mechanism's reachable workspace and confirm its flexibility, while the static equilibrium conditions verified its holding capacity. Physical experiments were also conducted to test the device's reliability in real-world scenarios. The static stability tests confirmed that the mechanism remains stable against gravity without drifting, and the force assessments showed that the actuation force is minimal, aligning well with the physical capabilities of average older adults.

The proposed mechanism offers a practical solution to the “Care Gap" and “Safety Gap" in current elderly care. Without the need for electronics, sensors, and waterproofing, this purely mechanical system ensures intrinsic safety in humid bathroom environments. It provides a low-cost, maintenance-free alternative to expensive robots or active mechatronic devices. Most importantly, it provided the ability for older adults to perform personal hygiene tasks independently, this technology has the potential to preserve the dignity and psychological well-being of the aging population.

Future work will focus on developing an active device to help older adults shower independently, potentially utilizing soft robotics technology. Additionally, we aim to integrate computer vision, deep learning, and reinforcement learning for adaptive, collision-free trajectory planning. This integration will enable multimodal, hands-free operation—such as gesture and voice control—allowing users to operate the assistive device fully autonomously with minimal physical exertion.

\bibliographystyle{asmems4}
\bibliography{asme2e}

\end{document}